\documentclass[letterpaper]{article} 
\usepackage{aaai23}  
\usepackage{times}  
\usepackage{helvet}  
\usepackage{courier}  
\usepackage[hyphens]{url}  
\usepackage{graphicx} 
\usepackage{natbib}  
\usepackage{caption} 
\usepackage{algorithm}
\usepackage{algorithmic}
\usepackage{times}  
\usepackage{tabularx}
\usepackage{rotating}
\usepackage{helvet}  
\usepackage{courier}  
\usepackage[hyphens]{url}  
\usepackage{graphicx} 
\usepackage{caption} 
\usepackage{algorithm}
\usepackage{algorithmic}

\usepackage{times}
\usepackage{latexsym}
\usepackage{graphicx}
\usepackage{pdfpages}
\graphicspath{ {./images/} }
\usepackage{bm}
\usepackage{svg}
\usepackage{amsmath}
\usepackage{amssymb}
\usepackage{booktabs}
\usepackage{multirow}
\usepackage{float}
\usepackage{xcolor,colortbl}
\usepackage{verbatim}
\usepackage{subcaption}
\usepackage{stfloats}
\usepackage{mathtools}
\usepackage{ulem}
\usepackage{array}

\newenvironment{shrinkeq}[1]
{ \bgroup
  \addtolength\abovedisplayshortskip{#1}
  \addtolength\abovedisplayskip{#1}
  \addtolength\belowdisplayshortskip{#1}
  \addtolength\belowdisplayskip{#1}}
{\egroup\ignorespacesafterend}

\usepackage{newfloat}
\usepackage{listings}
\DeclareCaptionStyle{ruled}{labelfont=normalfont,labelsep=colon,strut=off} 
\floatstyle{ruled}
\newfloat{listing}{tb}{lst}{}
\floatname{listing}{Listing}
\title{GraphPrompt: Graph-Based Prompt Templates for Biomedical Synonym Prediction}
\author{
    Hanwen Xu\equalcontrib\textsuperscript{\rm 1}, Jiayou Zhang\equalcontrib\textsuperscript{\rm 2}, Zhirui Wang\equalcontrib\textsuperscript{\rm 3}, Shizhuo Zhang\textsuperscript{\rm 4}, Megh Bhalerao\textsuperscript{\rm 1}, Yucong Liu\textsuperscript{\rm 5}, Dawei Zhu\textsuperscript{\rm 5}, Sheng Wang\thanks{Corresponding Author}\textsuperscript{\rm 1,}\textsuperscript{\rm 6}
}
\affiliations{
    \textsuperscript{\rm 1}University of Washington,\\
    \textsuperscript{\rm 2}Mohamed bin Zayed University of Artificial Intelligence\\
    \textsuperscript{\rm 3}Carnegie Mellon University,\\
    \textsuperscript{\rm 4}Nanyang Technological University, \\\textsuperscript{\rm 5}Peking University\\
    swang@cs.washington.edu

}

\begin{document}

\maketitle

\newcommand{\Enc}{\mathrm{Enc}}
\newcommand{\LQ}{``}
\newcommand{\CLS}{\mathrm{[CLS]}}
\newcommand{\RQ}{"}
\newcommand{\SEP}{\mathrm{[SEP]}}
\newcommand{\MASK}{\mathrm{[MASK]}}
\newcommand{\M}{\mathrm{[MASK]}}
\newcommand{\is}{\mathrm{is\ identical\ with}}
\newcommand{\isa}{\mathrm{is\ a\ kind\ of}}
\newcommand{\para}{\boldsymbol{\theta}}
\newcommand{\trans}{^\mathsf{T}}
\newcommand{\Temp}{\mathcal{T}}
\newcommand{\uu}{\boldsymbol{x}}
\newcommand{\vv}{\boldsymbol{y}}
\newcommand{\ww}{\boldsymbol{z}}
\newcommand{\argmax}[1]{\underset{#1}{\mathrm{argmax}}}
\newcommand{\sg}{\mathrm{sg}}
\newcommand{\NA}{\mathrm{null}}
\newcommand{\Red}{\mathrm{Red}}
\newcommand{\BioBERT}{\mathrm{BioBERT}}

\begin{abstract}
In the expansion of biomedical dataset, the same category may be labeled with different terms, thus being tedious and onerous to curate these terms. Therefore, automatically mapping synonymous terms onto the ontologies is desirable, which we name as biomedical synonym prediction task. Unlike biomedical concept normalization (BCN), no clues from context can be used to enhance synonym prediction, making it essential to extract graph features from ontology. We introduce an expert-curated dataset OBO-syn encompassing 70 different types of concepts and 2 million curated concept-term pairs for evaluating synonym prediction methods. We find BCN methods perform weakly on this task for not making full use of graph information. Therefore, we propose GraphPrompt, a prompt-based learning approach that creates prompt templates according to the graphs. GraphPrompt obtained 37.2\% and 28.5\% improvement on zero-shot and few-shot settings respectively, indicating the effectiveness of these graph-based prompt templates. We envision that our method GraphPrompt and OBO-syn dataset can be broadly applied to graph-based NLP tasks, and serve as the basis for analyzing diverse and accumulating biomedical data. All the data and codes are avalible at: https://github.com/HanwenXuTHU/GraphPrompt
\end{abstract}

\section{Introduction}

Mining biomedical text data, such as scientific literature and clinical notes, to generate hypotheses and validate discovery has led to many impactful clinical applications \citep{ zhao2021recent,lever2019cancermine}. One fundamental problem in biomedical text mining is entity synonym prediction, which aims to map a phrase to a concept in the controlled vocabulary \citep{sung2020biomedical}. Accurate entity synonym prediction enables us to summarize and compare biomedical insights across studies and obtain a holistic view of biomedical knowledge. Current approaches \citep{wright2019normco,ji2020bert,sung2020biomedical} to biomedical synonym prediction often focus on more standardized entities such as diseases \citep{dougan2014ncbi,li2016biocreative}, drugs \citep{kuhn2007stitch,pradhan2013task}, genes \citep{szklarczyk2016string} and adverse drug reactions \citep{roberts2017overview}. Despite their encouraging performance, these approaches have not yet been applied to the more ambiguous entities, such as processes, pathways, cellular components, and functions \citep{smith2007obo}, which lie at the center of life sciences. As scientists rely on these entities to describe disease and drug mechanisms \citep{yu2016translation}, the inconsistent terminology used across different labs inevitably hampers the scientific communication and collaboration, necessitating the synonym prediction of these entities.

\begin{figure}
\centering
\includegraphics[width=0.45\textwidth]{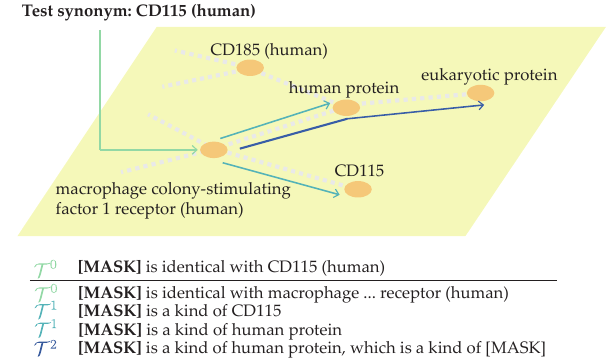}
\caption{Illustration of GraphPrompt. GraphPrompt classifies a test synonym (CD115 (human)) to an entity in the graph by converting the graph into prompt templates based on the zeroth-order neighbor ($\Temp^{0}$), first-order neighbors ($\Temp^{1}$), and second-order neighbors ($\Temp^{2}$).}
\label{figIntro}
\end{figure}

The first immediate bottleneck to achieve the normalization of these under-explored entities is the lack of a high-quality and large-scale dataset, which is the prerequisite for existing entity normalization approaches \citep{wright2019normco,ji2020bert,sung2020biomedical}. To tackle this problem, we collected 70 types of biomedical entities from OBO Foundry \cite{smith2007obo}, spanning a wide variety of biomedical areas and containing more than 2 million entity-synonym pairs. These pairs are all curated by domain experts and together form a high-quality and comprehensive controlled vocabulary for biomedical sciences, greatly augmenting existing biomedical synonym prediction datasets \citep{dougan2014ncbi,li2016biocreative,roberts2017overview}. The tedious and onerous curation of this high-quality dataset further confirms the necessity of developing data-driven approaches to automating this process and motivates us to introduce this dataset to the NLP community.

In addition to being the first large-scale dataset encompassing many under-explored entity types, this OBO-syn dataset presents a novel setting of graph-based synonym prediction. Specifically, entities of the same type form a relational directed acyclic graph (DAG), where each edge represents a relationship (e.g., $\mathrm{is\_a}$) between two entities. Intuitively, this DAG could assist the synonym prediction since nearby entities are biologically related, and thus more likely to be semantically and morphologically similar. Existing entity normalization and synonym prediction methods are incapable of considering the topological similarity from this rich graph structure \citep{wright2019normco,ji2020bert,sung2020biomedical}, limiting their performance, especially in the few-shot and zero-shot settings. Recently, prompt-based learning has demonstrated many successful NLP applications \citep{radford2019language,schick2020exploiting,jiang2020can}. The key idea of using prompt is to circumvent the requirement of a large number of labeled data by creating masked templates and then converting supervised learning tasks to a masked-language model task \cite{liu2021pre}. However, it remains unknown how to convert a large graph into text templates for prompt-based learning. Representing graphs as prompt templates might effectively integrate the topological similarity and textural similarity by alleviating the over-smoothing caused by propagating textual features on the graph. 

In this paper, we propose GraphPrompt, a prompt-based learning method for synonym prediction with the consideration of graph structures. The key idea of our method is to convert the graph structural information into prompt templates and solve a masked-language model task, rather than incorporating textual features into a graph-based framework. Our graph-based templates explicitly model the high-order neighbors (e.g., neighbors of neighbors) in the graph, which enables us to correctly classify synonyms that have relatively lower morphological similarity with the ground-truth entity (\textbf{Figure \ref{figIntro}}). Experiments on the novel OBO-syn dataset demonstrate the superior performance of our method against existing synonym prediction approaches, indicating the advantage of considering the graph structure. Case studies and the comparison to the conventional graph approach further reassure the effectiveness of our prompt templates, implicating opportunities on other graph-based NLP applications. Collectively, we introduce a novel biomedical synonym prediction task, a large-scale and high-quality dataset, and a novel prompt-based solution to advance biomedical synonym prediction. 

\section{Related Works}
\textbf{Biomedical synonym prediction.} Biomedical synonym prediction and entity normalization has been studied for decades because of its significance in a variety of biomedical applications. Conventional approaches mainly relied on rule-based methods \citep{d2015sieve,sullivan2011diego} or probabilistic graphical models \citep{leaman2013dnorm,leaman2016taggerone} to model the morphological similarity, which are incapable of normalizing functional entities that are semantically similar but morphologically different. Deep learning-based approaches \citep{li2017cnn,wright2019normco,pujary2020disease,deng2019ensemble,luo2018multi} and pre-trained language models (PLMs) \citep{ji2020bert, sung2020biomedical, lee2020biobert,miftahutdinov2021medical} have obtained encouraging results in capturing the semantics of entities through leveraging human annotations or large collections of corpus. However, these approaches focus on datasets comprising of less ambiguous entity types, such as drugs and diseases and are not able to incorporate graph structures into their framework. In contrast, we aim to utilize rich graph information to assist the synonym prediction of more ambiguous entities such as functions, pathways and processes.

\textbf{Incorporating graph structure into text modeling.} Graph-based approaches, such as network embedding \citep{tang2015line} and graph neural network \citep{kipf2016semi}, have been used to model the structural information in the text data, such as citation networks \citep{an2021enhancing}, social networks \citep{masood2021using,aljohani2020bot} and word dependency graph \citep{fu2019graphrel}. Among them, \citet{kotitsas2019embedding} considered the most similar DAG structure to our task and proposed a two-stage approach to integrate graph structure with textual information. The key difference between our method and existing approaches is that we transform the graph structures into prompt templates and then solve a masked-language model task, whereas existing works represent textual information as fixed node features and then optimize a graph-based model. 

\textbf{Prompt-based learning.}
Prompt-based learning have recently shown promising results in many applications\citep{liu2021pre}, such as text generation \citep{radford2019language,brown2020language}, text classification \citep{schick2020exploiting,gao2020making} and question answering \citep{khashabi2020unifiedqa,jiang2020can}. Prompt-based learning has not yet been applied to integrate the graph information. The most related prompt-based works to our task is prompt-based relation extraction \citep{chen2021adaprompt,han2021ptr}
and prompt-based knowledge base completion \citep{davison-etal-2019-commonsense}. These approaches only consider immediate neighbors in the graph and are not able to model more distant nodes, thus being incapable of capturing the topology of the entire graph. To the best of our knowledge, we are the first work that considers higher-order graph neighbors in the prompt-based learning framework.

\par
\section{Dataset Description and Analysis}

\begin{figure*}
\centering
\includegraphics[width=0.9\textwidth]{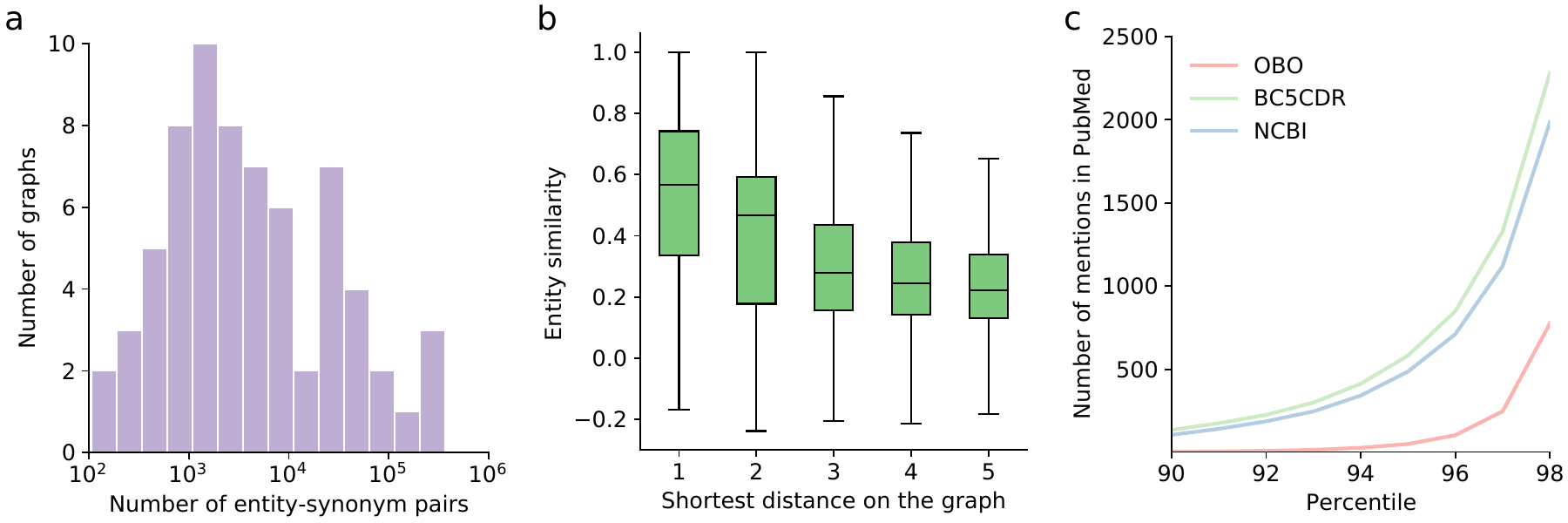}
\caption{Analysis of the OBO-syn dataset. a, Bar plot showing the distribution of the number of entity-synonym pairs in 70 graphs. b, Box plot comparing the textual similarity of entity pairs having different shortest distances on the graph. c, Line chart comparing the phrase mentions of NCBI, BC5CDR, OBO-syn. The y-axis is the number of mentions in 28 million PubMed abstracts. The x-axis is the phrase percentile sorted by the number of mentions. }
\label{figData}
\end{figure*}

We collected 70 relational graphs from Open Biological and Biomedical Ontology Foundry (OBO) \citep{smith2007obo}. Nodes in the same relational graph represent biomedical entities belonging to the same type, such as protein functions, cell types, and disease pathways. Each edge represents a relational type, such as $\mathrm{is\_a}$, $\mathrm{part\_of}$, $\mathrm{capable\_of}$, and $\mathrm{regulates}$. We leveraged these edge types to build templates in our prompt-based learning framework. The number of nodes in each graph ranges from 113 to 2,334,910 with a median value of 3,077. The number of synonyms for each entity ranges from 1 to 284 with a median value of 2 (ignoring the entities without synonyms). On average, each graph has 34,418 entity-synonym pairs and 72.9\% of graphs have more than 1,000 entity-synonym pairs (\textbf{Figure \ref{figData}a}). The graph structure and entity synonym associations are all curated by domain experts, presenting a large-scale and high-quality collection.

In comparison to other biomedical entity synonym prediction datasets \citep{dougan2014ncbi,li2016biocreative,roberts2017overview}, OBO-syn presents a unique graph structure among entities. Intuitively, nearby entities as well as their synonyms should be semantically similar, as their biological concepts are relevant. To validate this intuition, we investigated the consistency between graph-based entity similarity and text-based entity similarity. In particular, we used the shortest distance on the graph to calculate graph-based similarity and Sentence-BERT \cite{reimers2019sentence} to calculate text-based similarity. We observed a strong correlation between these two similarity scores (\textbf{Figure \ref{figData}b}), suggesting the possibility to transfer synonym annotations from nearby entities to improve the synonym prediction. 

We next compared this OBO-syn dataset with the existing biomedical entity normalization dataset. We first observed very small overlaps of 5.26\%, 14.59\%, 3.29\%  between our dataset and three widely-used biomedical synonym prediction datasets NCBI-disease \citep{dougan2014ncbi}, BC5CDR-disease \citep{li2016biocreative}, and BC5CDR-chemical \citep{li2016biocreative}, respectively. The small overlaps with existing datasets indicate the uniqueness of our dataset, and further make us question the performance of the state-of-the-art synonym prediction methods on this novel dataset. More importantly, we noticed a substantially large number of out-of-vocabulary phrases in our dataset compared to existing datasets (\textbf{Figure \ref{figData}c}). We calculate the number of mentions of each phrase in 29 million PubMed abstracts, which are used as the pre-training corpus for biomedical pre-trained models \citep{lee2020biobert,gu2020domain}. The 95 percentile of the number of mentions in our dataset is only 51, substantially lower than 487 in NCBI and 582 in BC5CDR, suggesting a worse generalization ability using pre-trained language models and motivating us to exploit the graph structures for this dataset.

\section{Problem Statement}

The goal of entity synonym prediction is to map a given synonym phrase $s$ to the corresponding entity $v$ based on their semantic similarity. One unique feature of our problem setting is that entities belonging to the same type form a relational graph. Formally, we denote this relational graph as $\mathcal{G}=(\mathcal{V}, \mathcal{R}, \mathcal{E})$, where $\mathcal{V}$ is the set of entities, $\mathcal{R}$ is the set of relation types and $\mathcal{E}\subset \mathcal{V} \times \mathcal{R} \times \mathcal{V}$ is the set of edges. Let $\mathcal{C}$ be the vocabulary of the corpus. Each node $v_i \in \mathcal{V}$ is represented as an entity phrase $v_i \triangleq \langle v_i^1,v_i^2,\dots,v_i^{|v_i|} \rangle$, where $v_i^k \in \mathcal{C}$. In addition to the graph, we also have a set of mapped synonyms $\mathcal{S}$ that will be used as the training data. Each  $s_j \triangleq \langle s_j^1,s_j^2,\dots,s_j^{|s_j|} \rangle \in \mathcal{S}$ is mapped to one entity $v_i$ in the graph $\mathcal{S}$, and $s_j^k \in \mathcal{C}$.

Our goal is to classify a test synonym $s'$ to an entity $v$ in the graph. Since the majority of entities only have very few synonyms (e.g., 96.9\% of entities have less than 5 synonyms), we consider a few-shot and a zero-shot setting. Specifically, in the few-shot setting, the test set entities are included in the training set entities. On the contrary, the entities in the training set and the test set present no overlap in the zero-shot setting, and therefore the entities of training datasets are unobservable for test procedure. The small number of training synonyms for each entity could exacerbate over-fitting. To mitigate the over-fitting problem, we propose graph-based prompt templates, where we consider the synonyms of nearby entities in the training data.

\subsection{Base Model}
We first introduce a base model that only considers the textual information of synonyms and entities while disregarding the graph structure. Following the previous work \citep{sung2020biomedical}, the base model uses two encoders to calculate the similarity between the queried synonym $s$ and the candidate entity $v$. The first encoder $\Enc_s$ encodes the queried synonym into the dense representation $\uu_s=\Enc_s(t_s)$. The second encoder $\Enc_v$ encodes the candidate entity into the dense representation $\uu_v=\Enc_v(t_v)$. Then the predicted probability of choosing entity $v$ is calculated as $P(\uu_v|\uu_s)=\frac{Q(\uu_v, \uu_s)}{\sum_{v'\in \mathcal{V}} Q(\uu_{v'}, \uu_s)}$,
where $Q$ is defined as $Q(\uu_v, \uu_s)=\exp(\uu_v \trans \uu_s)$. We select BioBERT with $\CLS$ readout function as $\Enc_v$ and $\Enc_s$, and share the parameters between both encoders. Following \citet{sung2020biomedical}, the input $t_v$ and $t_s$ are designed as $\LQ\CLS \ v\ \SEP\RQ$ and $\LQ\CLS\ s\ \SEP\RQ$ respectively. In practice, we find that the initial $\CLS$ output vectors are fairly close. This can result in large positive $\uu_v \trans\uu_s$, which leads to slow convergence and potential numerical issues, yet it is not addressed by BioSyn \citep{sung2020biomedical}. To alleviate this issue, we use a trainable 1-d BatchNorm layer and redefine our similarity function $Q$ as: 
\begin{shrinkeq}{-1ex} \begin{equation}
    Q(\uu_v, \uu_s)=\exp(\mathrm{BN}(\uu_v \trans \uu_s)). \label{simMN}
\end{equation} \end{shrinkeq}

When the candidate entity set is large, back-propagating through $\uu_v$ results in high memory complexity due to the construction of $|\mathcal{V}|$ computation graphs to get $\uu_{v'}$. To tackle this problem, we apply the stop gradient trick to $\uu_{v'}$, following \citet{sung2020biomedical}. Besides, we utilize the \textit{hard negative} strategy following \citet{sung2020biomedical} by sampling difficult negative candidates $\mathcal{U} \subset \mathcal{V}$. The loss function is defined as:
\begin{shrinkeq}{-1ex}\begin{equation}
    \mathcal{L}_{\mathrm{base}} = - \sum_{(v,s)} \log  \frac{Q(\sg(\uu_v), \uu_s)}{\sum_{v'\in \mathcal{U}\cup \{v\}} Q(\sg(\uu_{v'}), \uu_s)},
\end{equation}\end{shrinkeq}
where $\sg$ denotes the stop gradient operation.
Besides, To further save computation time, we cache the values of $\sg(\uu_{v'}), v'\in \mathcal{V}$ and iteratively update them.  

\subsection{Prompt Model}
Previous work \citep{sung2020biomedical} and the base model use BioBERT with $\CLS$ readout function as the encoders, which take the synonym or entity as the input and use the hidden state of $\CLS$ as the output. However, using the synonym or entity as the input text might not fully capture its semantic since PLMs are often pre-trained with sentences instead of phrases. To tackle this problem, we construct two simple prompt templates $\Temp^0$ for a training  entity-synonym pair $(v, s)$ as:
$\Temp^0(\MASK, v) =  \LQ \MASK$ is identical with $v \RQ$
and
$\Temp^0(\MASK, s) =  \LQ \MASK$ is identical with $s \RQ$ ($\CLS$ and $\SEP$ are omitted).
Then we optimize the model by solving an masked language modeling task, where we use the output of BioBERT at $\MASK$ token as the dense representation $\uu_v$ ($\uu_s$) for $v$ ($s$), respectively: $\uu_v = \BioBERT (\Temp^0(\MASK, v)),
     \uu_s = \BioBERT (\Temp^0(\MASK, s)). \label{originU}$

Since the graph is not used here, we refer to $\uu_v$ as the zeroth-order representation of entity $v$. The loss function of prompt model is similar to base model's, where we select the whole entity set $\mathcal{V}$ as candidates instead of its subset:

\begin{shrinkeq}{-1ex}\begin{equation}
    \mathcal{L}_{\mathrm{p}} = - \sum_{(v,s)} \log  \frac{Q(\sg(\uu_v), \uu_s)}{\sum_{v'\in \mathcal{V}} Q(\sg(\uu_{v'}), \uu_s)}.
\end{equation}\end{shrinkeq}


\section{GraphPrompt Model}
\subsection{Intuition}
The observation that nearby entities are more semantically similar (\textbf{Figure \ref{figData}b}) motivates us to integrate textual similarity with graph topological similarity to boost the entity normalization. Conventional approaches often integrate text and graph information by adapting a graph-based framework and incorporating text features as node features \cite{kotitsas2019embedding}. However, such approaches might not fully utilize the strong generalization ability of pre-trained models, which have been crucial for a variety of NLP tasks \citep{devlin2018bert,petroni2019language}. In contrast to conventional approaches, we propose to utilize a prompt-based learning framework to integrate text and graph information through representing the graph information as prompt templates. To the best of our knowledge, our method is the first attempt to represent the graph structure as prompt templates. 

\subsection{First-order GraphPrompt}
GraphPrompt utilizes the graph information during training. GraphPrompt considers first-order neighborhood (i.e., immediate neighbors) and second-order neighborhood (i.e., neighbors of the neighbors) to construct prompt templates for a given entity.

To model first-order neighbors, GraphPrompt defines the template $ \Temp_r^1(v_i, v_j) = \LQ v_i\ r' \  v_j\RQ$ for an edge between entity $v_i$ and its immediate neighbor entity $v_j$ with relation type $r$. $r'$ is created from $r$ with minor morphological change. For a given triple $(v_i, r, v_j)$ in the graph, we create a masked-language model task by randomly masking $v_i$ or $v_j$. We also include the template that replaces the unmasked $v$ with its training synonym $s$. For example, when $v_i$ is masked and $v_j$ is replaced with $s_k$, we obtain the following template: $\Temp_r^1(\M , v_j) = \LQ \MASK \ r' \  s_k\RQ$. We then use BioBERT to obtain the first-order representation $\vv_{v_i}$ based on this template:
\begin{shrinkeq}{-1ex}\begin{equation}
    \vv_{v_i} = \BioBERT (\Temp_r^1(\M , v_j)).
\end{equation}\end{shrinkeq}

We then calculate the loss term by comparing the first-order representation of $v_i$ with the zeroth-order presentation $\uu_{v_i}$:

\begin{shrinkeq}{-1ex}
\begin{gather}
    \mathcal{L}_1 = - \sum_{(v_i, v_j)}\log P(\uu_{v_i}|\vv_{v_i})  \label{L1}
\end{gather}
\end{shrinkeq}
\begin{table*}[ht]
\centering
{
\fontsize{9pt}{3pt}\selectfont
\setlength{\tabcolsep}{2pt}
    \begin{tabular}{@{}lcccccccccccc@{}}
\toprule
Dataset                      & \multicolumn{4}{c}{mp}                                                                                                                                            & \multicolumn{4}{c}{cl}                                                                                                                                            & \multicolumn{4}{c}{hp}  \\

data split                   & \multicolumn{2}{c}{zero-shot}                                                   & \multicolumn{2}{c}{few-shot}                                                    & \multicolumn{2}{c}{zero-shot}                                                   & \multicolumn{2}{c}{few-shot}                                                    & \multicolumn{2}{c}{zero-shot}                                                   & \multicolumn{2}{c}{few-shot}                                                                                \\
                             & Acc@1                                  & Acc@10                                 & Acc@1                                  & Acc@10                                 & Acc@1                                  & Acc@10                                 & Acc@1                                  & Acc@10                                 & Acc@1                                  & Acc@10                                 & Acc@1                                  & Acc@10                       \\
\midrule
Sieve-Based                  & 3.40                                   & --                                     & --                                     & --                                     & 6.80                                   & --                                     & --                                     & --                                     & 4.60                                   & --                                     & --                                     & --                                      \\
BNE                          & 43.60                                  & 68.10                                  & 51.70                                  & 70.90                                  & 40.00                                  & 65.30                                  & 49.20                                  & 66.60                                  & 32.50                                  & 58.40                                  & 36.70                                  & 59.40                                \\
NormCo                       & --                                     & --                                     & 41.25                                  & 53.48                                  & --                                     & --                                     & 52.68                                  & 59.76                                  & --                                     & --                                     & 49.44                                  & 55.25                                \\
TripletNet                   & 40.67                                  & 67.54                                  & 42.39                                  & 68.75                                  & 28.81                                  & 60.07                                  & 28.61                                  & 60.96                                  & 26.54                                  & 51.99                                  & 27.69                                  & 52.29                             \\
BioSyn                       & 62.04                                  & 74.08                                  & 73.55                                  & 82.67                                  & 53.55                                  & 66.18                                  & 64.34                                  & 75.59                                  & 50.72                                  & 67.24                                  & 59.84                                  & 72.25                             \\
GCN                          & 70.88                                  & 88.61                                  & 83.83                                  & 93.18                                  & 60.49                                  & 84.09                                  & 74.95                                  & 90.50                                  & 54.92                                  & 79.47                                  & 68.39                                  & 87.17                           \\
\midrule
Base model                   & 76.47                                  & 88.38                                  & 85.78                                  & 93.02                                  & 65.14                                  & 82.58                                  & 76.64                                  & 89.19                                  & 61.53                                  & 78.14                                  & 72.04                                  & 86.38                             \\
Prompt                       & 79.51          & 89.74                                  & 87.56                                  & 95.41                                  & 66.45                                  & 82.02                                  & 80.93                                  & 93.33                                  & 62.80                                  & 81.67                                  & 75.12                                  & 90.40                                  \\
GraphPrompt ( w/o $\Temp^2$) & 79.40                                  & 90.95          & 88.05          & \textbf{95.75} & \textbf{68.00}          & 84.89 & 82.04          & 93.81          & 66.36          & 84.24          & 77.60          & 92.33          \\
\midrule
GraphPrompt                  & \textbf{81.14} & \textbf{92.90} & \textbf{88.86} & 95.61          & 67.81 & \textbf{87.68}          & \textbf{82.86} & \textbf{94.24} & \textbf{69.14} & \textbf{88.07} & \textbf{77.87} & \textbf{92.66}\\

Dataset                                                                                                                                                & \multicolumn{4}{c}{fbbt}                                                                                                                                          & \multicolumn{4}{c}{doid} & \multicolumn{4}{c}{}                                                                                                                                        \\
data split                                       & \multicolumn{2}{c}{zero-shot}                                                   & \multicolumn{2}{c}{few-shot}                                                    & \multicolumn{2}{c}{zero-shot}                                                   & \multicolumn{2}{c}{few-shot}                                                    \\
                            & Acc@1                                  & Acc@10                                 & Acc@1                                  & Acc@10                                 & Acc@1                                  & Acc@10                                 & Acc@1                                  & Acc@10                                 \\
\midrule
Sieve-Based                                & 1.00                                   & --                                     & --                                     & --                                     & 8.20                                   & --                                     & --                                     & --                                     \\
BNE                                       & 34.30                                  & 59.80                                  & 44.40                                  & 62.10                                  & 40.10                                  & 58.40                                  & 40.00                                  & 57.40                                  \\
NormCo                                                      & --                                     & --                                     & 32.58                                  & 42.57                                  & --                                     & --                                     & 58.23                                  & 67.01                                  \\
TripletNet                        & 6.21                                   & 14.14                                  & 3.15                                   & 9.46                                   & 26.59                                  & 43.34                                  & 25.79                                  & 42.92                                  \\
BioSyn                                                 & 40.38                                  & 61.95                                  & 57.32                                  & 68.06                                  & 40.68                                  & 53.82                                  & 48.90                                  & 61.04                                  \\
GCN                                       & 48.98                                  & 73.68                                  & 54.91                                  & 78.50                                  & 46.57                                  & 65.91                                  & 55.27                                  & 75.42                                  \\
\midrule
Base model                                    & 65.38                                  & 79.24                                  & 69.76                                  & 84.08                                  & 50.71                                  & 64.86                                  & 59.40                                  & 74.69                                  \\
Prompt                           & 68.53          & 79.29                                  & 73.75                                  & 86.49                                  & 56.43                                  & 69.51                                  & 66.31                                  & 81.03                                  \\
GraphPrompt ( w/o $\Temp^2$)    & 68.10                                  & 80.08          & \textbf{75.65} & 89.06          & 56.85          & 70.67 & 66.70          & 81.99          \\
\midrule
GraphPrompt                & \textbf{69.01} & \textbf{84.54} & 75.55          & \textbf{89.74} & \textbf{59.06} & \textbf{73.78} & \textbf{67.53} & \textbf{82.50}  \\
\bottomrule
\end{tabular}
}
\caption{\label{table1:allresults} The performance of our method and comparison approaches on 5 datasets using zero-shot and few-shot settings. The best model in each column is colored in blue and the second best is colored in light blue.}
\end{table*}

\subsection{Second-order GraphPrompt}
To consider second-order neighbors, GraphPrompt first finds all 2-hop relational paths $(v_i, r, v_j, \mathrm{is\_a}, v_k)$ in the graph. Since $\mathrm{is\_a}$ relation contributes to the majority of the relation type, we fix the second relation to be $\mathrm{is\_a}$ for simplicity. The prompt template is then defined as $ \Temp^2_r(v_i, v_j, v_k) = \LQ v_i\ r' \  v_j,\ \mathrm{which}\ \isa\ v_k\RQ$.

Different from $\Temp^0$ and $\Temp^1$, there are three tokens that can be masked in $\Temp^2$. We chose to mask two tokens in each template, resulting in two kinds of second-order templates: 
\begin{small}
\begin{shrinkeq}{-1ex}\begin{gather*}
		 \ww_{v_i}, \ww_{v_k}=\BioBERT(\Temp_r^2(\M, v_j, \M)) \\
		 \ww_{v_j}, \ww_{v_k}=\BioBERT(\Temp_r^2(v_i, \M, \M))
\end{gather*}\end{shrinkeq}
\end{small}
We don't consider the template of $\Temp_r^2(\M, \M, v_k)$ because of the DAG structure in our dataset. The numbers of child nodes and grandchild nodes grow exponentially in DAG and will introduce too many paths using $\Temp_r^2(\M, \M, v_k)$ template, slowing down the optimization. 

To calculate the loss term based on $\Temp_r^2(\M, v_j, \M)$, we compare the second-order dense representation $\ww_{v_i}$, $\ww_{v_k}$ to the zeroth-order dense representation $\uu_{v_i}$, $\uu_{v_k}$. $\ww_{v_j}'$, $\ww_{v_k}'$ and the loss term based on $\Temp_r^2(v_i, \M, \M)$ is defined similarly. We define the loss term for second-order neighbors as $\mathcal{L}_2 = - \sum_{(v_i, v_j, v_k)} (\log P(\uu_{v_i}|\ww_{v_i}) + \log P(\uu_{v_k}|\ww_{v_k}) \\
     + \log P(\uu_{v_j}|\ww_{v_j}') + \log P(\uu_{v_k}|\ww_{v_k}'))$. Although we can further define higher-order templates accordingly, we observed limited improvement by including third-order or even higher-order templates in our experiments. This observation is consistent with conventional graph embedding approaches where only first-order and second-order neighborhood are explicitly modeled \cite{tang2015line}. For the 2-hop relational path, we didn't consider sibling-based templates such as $\LQ$Both $\M$ and $\M$ are a kind of $v\RQ$ due to the large number of sibling pairs in the DAG. Nevertheless, such templates might be worth exploring on other graphs.

In practice, different entities may have similar $\uu_{v}$, making them indistinguishable at the test stage. 
To alleviate this problem, we hope the embeddings of different entities to be different, except for the neighbor nodes on the graph. Therefore we consider another contrastive loss term $\mathcal{L}_\mathrm{c}$ that first encourages the model to distinguish different entities and second increases the similarity of neighbor entities $(v_{i}, r, v_{j})$. The contrastive loss considers $\mathcal{L}_{\mathrm{c_{1}}} = - \sum_{v} \log P(\uu_v|\uu_v)$ and $\mathcal{L}_{\mathrm{c_{2}}} = - \sum_{(v_{i}, v_{j})} \log P(\uu_{v_{i}}|\uu_{v_{j}})$ and the final loss is $\mathcal{L}_{\mathrm{c}} = \lambda_\mathrm{c_{1}} \mathcal{L}_\mathrm{c_{1}} + \lambda_\mathrm{c_{2}} \mathcal{L}_\mathrm{c_{2}}$.
The final loss of our model combines of $\mathcal{L}_{\mathrm{p}}$, $\mathcal{L}_{\mathrm{c_{1}}}$,$\mathcal{L}_{\mathrm{c_{2}}}$, $\mathcal{L}_{1}$ and $\mathcal{L}_{2}$, with weights $\lambda_{\mathrm{p}}$, $\lambda_{c_{1}}$, $\lambda_c{_{2}}$, $\lambda_1$ and $\lambda_2$ chosen on the validation set:
\begin{shrinkeq}{-1ex}\begin{equation}
    \mathcal{L} = \lambda_{\mathrm{p}} \mathcal{L}_{\mathrm{p}} + \lambda_\mathrm{c_{1}} \mathcal{L}_\mathrm{c_{1}} + \lambda_\mathrm{c_{2}} \mathcal{L}_\mathrm{c_{2}} + \lambda_1 \mathcal{L}_1 + \lambda_2 \mathcal{L}_2
\end{equation}\end{shrinkeq}

The possible choices of these weights are 2, 3 for $\mathcal{L}_{\mathrm{p}}$, 0 and 1 for $\mathcal{L}_{\mathrm{c_{1}}}$, 0 and 1 for $\mathcal{L}_{\mathrm{c_{2}}}$, 1 and 2 for $\lambda_1$ and $\lambda_2$. We used grid search to find the best combinations of these values. In our implementations, we found $\mathcal{L}_{\mathrm{c_{2}}}$ could obtain higher accuracy in the zero-shot setting but the performance deteriorated in the few-shot setting. Therefore we set $\lambda_\mathrm{c_{2}}$ to 1 in zero-shot setting and 0 in the few-shot setting.
\begin{figure*}
\centering
\includegraphics[width=0.9\textwidth]{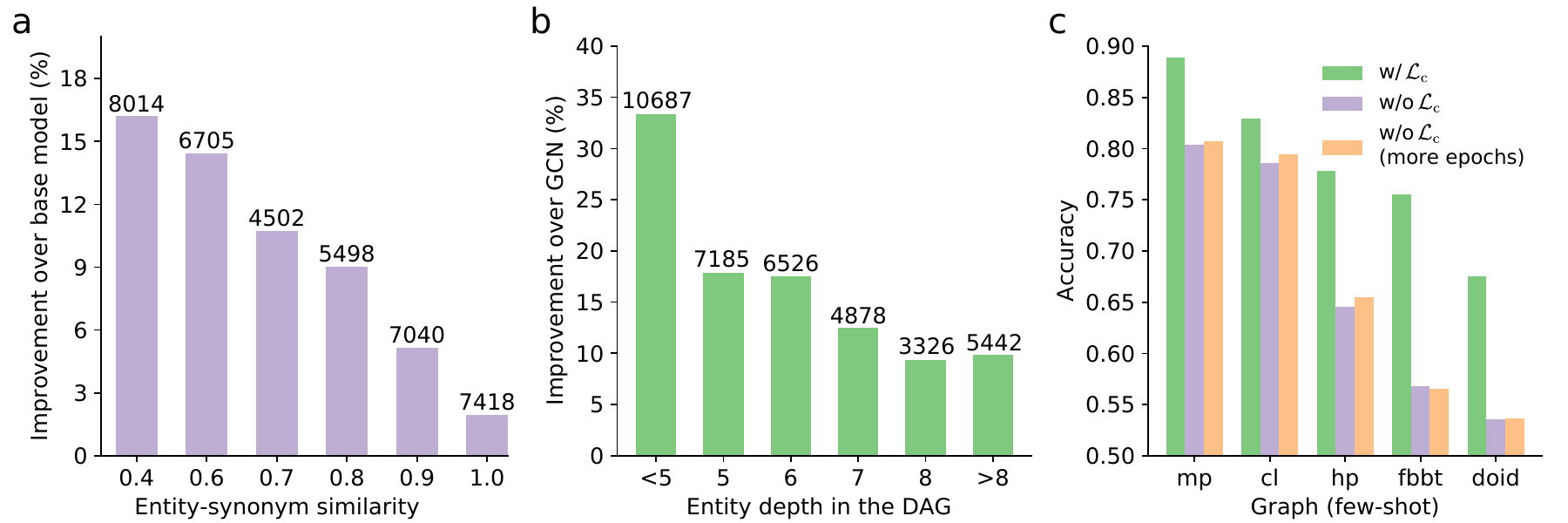}
\caption{Performance analysis of GraphPrompt. a, Bar plot showing the improvement of GraphPrompt over the base model under different entity-synonym similarity intervals. x-axis is the upper bound of the interval (e.g., 0.6 stands for [0.4-0.6]). b, Bar plot showing the improvement of GraphPrompt over GCN under different entity depths in the DAG. c, Bar plot showing the effect of $\mathcal{L}_{\mathrm{c}}$ in the few-shot setting. Orange bar stands for training 2 times more epochs without using $\mathcal{L}_{\mathrm{c}}$.}
\label{figExp}
\end{figure*}
\section{Experimental Results}
\subsection{Experimental Settings}
We selected five graphs (mp, cl, hp, fbbt, doid) with the number of entities between 10,000 and 20,000 from OBO-syn. We investigated a few-shot setting and a zero-shot setting. In the few-shot setting, we split the synonyms into six folds, and then used four folds as training set, one fold as validation set and one fold as test set. In the zero-shot setting, we split all entities into three folds, and then used two folds as training set and one fold as test set. All synonyms of training (test) entities are observable (unobservable) during training. Our method and all comparison approaches used the same data split.

We compared our method to the state-of-the-art synonym prediction approaches: Sieve-Based \citep{d2015sieve} , BNE \citep{phan2019robust}, NormCo \citep{wright2019normco}, TripletNet \citep{mondal2020medical} and BioSyn \citep{sung2020biomedical}, and a graph convolutional network (GCN) \citep{kipf2016semi}. We also compared our method with a base model ($\mathcal{L}_{base}$), a prompt model ($\mathcal{L}_\mathrm{p} + \mathcal{L}_\mathrm{c}$) and a first-order GraphPrompt (w / o $\Temp^2$) ($\mathcal{L}_1 + \mathcal{L}_\mathrm{p} + \mathcal{L}_\mathrm{c}$). 


\subsection{Improved Performance in Few-shot Setting}
We first sought to evaluate the performance of our method in the few-shot setting (\textbf{Table \ref{table1:allresults}}). We found that our method outperformed all other approaches in all metrics on all the datasets. When comparing to the best-performed synonym prediction approach BioSyn, our method obtains an average 27.7\% improvement on Acc@10 and 35.5\% improvement on Acc@1, indicating the prominence of using the graph structure to leverage annotations from nearby entities. We found that using graph structure leads to large improvement on datasets with fewer training samples (39.6\% improvement on doid comparing to 24.9\% on mp), suggesting GraphPrompt's ability to learn from limited samples.

We next compared our method to a graph-based approach GCN and observed a superior performance of GraphPrompt, confirming the effectiveness of modeling graph structures using prompt templates. The base model, which does not exploit the graph structure, also performed better than GCN, partially due to the over-smoothing issue in GCN. Despite showing a less superior performance comparing to our method, GCN still outperformed most of the synonym prediction approaches that do not consider graph structure, reassuring the advantage of using graph structure in this dataset. 

To further verify that the improvement of our method comes from using graph structure, we compared the performance of GraphPrompt with the base prompt model and the first-order prompt model. Overall, GraphPrompt is better than both approaches by utilizing the second-order neighborhood, while the first-order prompt is better than the base prompt model. Collectively, our results clearly assure the importance of considering the graph structure and the effectiveness of modeling it using prompt templates.
\subsection{Improved Performance in Zero-shot Setting}
After verifying the superior performance of our method in few-shot learning, we next investigate the more challenging zero-shot setting, where ground-truth entities in the test set have no synonyms in the training set (\textbf{Table \ref{table1:allresults}}). Likewise, our method outperformed all comparison approaches in all metrics on all datasets. We found that GraphPrompt obtained larger improvement over BioSyn in the zero-shot setting compared to the few-shot setting. Since ground-truth entities do not have any observed synonyms in the zero-shot setting, graph information becomes more crucial to aggregate synonym annotations from nearby entities.

The consistent improvement of GraphPrompt over GCN in both zero-shot and few-shot settings further confirms the effectiveness of using prompt templates to capture the graph structure. GraphPrompt also shows consistent improvement over the base prompt model and the first-order prompt model, indicating the importance of considering second-order neighbors in the graph.


\subsection{Improvement Analysis}
We sought to investigate the superior performance of GraphPrompt. We first calculated the textual similarity between the test synonyms and their ground truth entities using Sentence-BERT \citep{reimers2019sentence}. We found that the improvement of GraphPrompt over the base model increases with the decreasing of this textual similarity (\textbf{Figure \ref{figExp}a}). Entity-synonym pairs that have smaller textual similarity are more difficult to be predicted correctly with only the textual information, thus obtaining larger improvement from the graph structure. Moreover, the low overlaps with pre-training corpus limit the knowledge from PLMs, necessitating the consideration of graph information. 

We then sought to study the improvement of GraphPrompt over GCN. Interestingly, we found that GCN tends to have better performance on Acc@10 rather than Acc@1, whereas our method shows consistent improvement on these two metrics. As GCN is known to suffer from over-smoothing \citep{li2018deeper, hoang2019revisiting}, it might not distinguish very close entities on the graph, leading to much worse top 1 prediction performance, but better top 10 prediction performance. In contrast, our prompt-based graph learning does not show a performance degrade in top 1 prediction, suggesting that our method is less prone to over-smoothing. 

\begin{table*}[htb]
\centering
{
\fontsize{9pt}{9pt}\selectfont
\setlength{\tabcolsep}{0.3pt}
\begin{tabular}{lll}
\hline
\textbf{Test synonym}             & \multicolumn{2}{l}{adult {Leucokinin ABLK neuron of the abdominal ganglion}}  \\
\textbf{Ground-truth entity}              & \multicolumn{2}{l}{\uwave{adult abdominal ganglion Leucokinin neuron}} \\ 
\textbf{Baseline predictions}               & \multicolumn{2}{l}{adult Leucokinin neuron of the central nervous system (\textbf{GCN}), adult anterior LK Leucokinin neuron (\textbf{Base model})}                                                                              \\ \hline
\textbf{Prompt templates}          & \multicolumn{2}{l}{\begin{tabular}[c]{@{}l@{}}{[}\uwave{adult abdominal ganglion Leucokinin neuron}{]} is a kind of \underline{abdominal neuron}, which is a kind of {[}...{]}\\ 
{[}...{]} is identical with larval {Leucokinin ABLK neuron of the abdominal ganglion}, which is a kind of {[}\underline{abdominal neuron}{]}\end{tabular}} \\ \hline
\end{tabular}
}
\caption{An example of GraphPrompt prediction. Selected second-order prompt templates that affect the results are listed. Masked tokens are displayed within brackets.} \label{case2}
\end{table*}

To further verify this, we examined the improvement of our method against GCN at different depths in the graph (\textbf{Figure \ref{figExp}b}). We found that the improvement of our method over GCN becomes larger when the depth of the entity is smaller. Because of the DAG structure in our graph, entities that have smaller depth are closer to the center of the graph, and could be more disturbed by the over-smoothing issue. In contrast, our method explicitly converts the graph structure into prompt templates, successfully alleviating the over-smoothing issue caused by propagating on the entire graph.

Next, we examined the effect of the $\mathcal{L}_\mathrm{c}$ norm in our method (\textbf{Figure \ref{figExp}c}). As expected, adding $\mathcal{L}_\mathrm{c}$ greatly improved the performance on all the datasets in the few-shot setting. The improvement is much larger on datasets that have worse overall performance (e.g., fbbt, doid), indicating the importance of separating the embeddings of different entities. We also noticed that the accuracy of the state-of-the-art entity normalization approaches, such as BioSyn and NormCo, is much worse on our OBO-syn dataset than on the mainstream datasets, such as BC5CDR and NCBI (see results in \citet{sung2020biomedical} ), further confirming the difficulty of our task and dataset.

Finally, we presented two case studies of how GraphPrompt utilized the graph structure to correctly identify the  entity (\textbf{Figure \ref{figIntro}} and \textbf{Table \ref{case2}}). We found that GraphPrompt performed a `recombination' of two nearby phrases using the graph-based prompt templates during the prediction. For example, GraphPrompt correctly classified the test synonym `adult Leucokinin ABLK neuron of the abdominal ganglion' to the entity `adult abdominal ganglion Leucokinin neuron' by combining it with the second-order neighbor `larval Leucokinin ABLK neuron of the abdominal ganglion', whereas comparison approaches classified to incorrect but semantically similar entities (e.g., `adult anterior LK Leucokinin neuron') (\textbf{Table \ref{case2}}).  Likewise, GraphPrompt correctly classified `CD115 (human)' to `macrophage ... receptor (human)' by recombining it with CD115 according to the first-order prompt template. These recombinations of nearby entities reassure the effectiveness of graph-based prompts in biomedical entity normalization. 

\section{Conclusion and Future Work}
We have presented a novel biomedical synonym prediction dataset OBO-syn that encompasses 70 biomedical entity types and 2 million entity-synonym pairs. OBO-syn has demonstrated small overlaps with existing datasets and more challenging entity-synonym predictions. To leverage the unique graph structures in OBO-syn, we have proposed GraphPrompt, which converts graph structures into prompt templates and then solves a masked-language model task. GraphPrompt has obtained superior performance to the state-of-the-art synonym prediction approaches on both few-shot and zero-shot settings. 

In spite of the superior performance of GraphPrompt, there are still few limitations to tackle in the future work. First, GraphPrompt requires the entity to be within one predefined graph, which may hinder the applications on novel entities that are not on the graph. Therefore we plan to supplement the entity with their textual description, such as the "def" field in the OBO. Clues from the textual description can further restore the missing links between the novel entities and existing entities. Second, the text describing relations between entities in our study might be simple, which limits the further improvement of GraphPrompt and applications on more complex networks. We can utilize literature mining to incorporate more details in the relation description texts.

Since GraphPrompt can in principle be applied to integrate other types of graphs and text information, we are interested in exploiting GraphPrompt in other graph-based NLP tasks, such as citation network analysis and graph-based text generation. The novel OBO-syn dataset can also advance tasks beyond entity normalization, such as link prediction, graph representation learning, and be integrated with other scientific literature datasets to investigate entity linking, key phrase mining, and named entity recognition. We envision that our method GraphPrompt and OBO-syn will pave the path for comprehensively analyzing diverse and accumulating biomedical data. 

\bibliography{aaai23}



\end{document}


\newcommand{\Enc}{\mathrm{Enc}}
\newcommand{\LQ}{``}
\newcommand{\CLS}{\mathrm{[CLS]}}
\newcommand{\RQ}{"}
\newcommand{\SEP}{\mathrm{[SEP]}}
\newcommand{\MASK}{\mathrm{[MASK]}}
\newcommand{\M}{\mathrm{[MASK]}}
\newcommand{\is}{\mathrm{is\ identical\ with}}
\newcommand{\isa}{\mathrm{is\ a\ kind\ of}}
\newcommand{\para}{\boldsymbol{\theta}}
\newcommand{\trans}{^\mathsf{T}}
\newcommand{\Temp}{\mathcal{T}}
\newcommand{\uu}{\boldsymbol{x}}
\newcommand{\vv}{\boldsymbol{y}}
\newcommand{\ww}{\boldsymbol{z}}
\newcommand{\argmax}[1]{\underset{#1}{\mathrm{argmax}}}
\newcommand{\sg}{\mathrm{sg}}
\newcommand{\NA}{\mathrm{null}}
\newcommand{\Red}{\mathrm{Red}}
\newcommand{\BioBERT}{\mathrm{BioBERT}}

\appendix
\section{Appendix}
\subsection{Relations and phrases}
\label{subsec:appendixA1}
Table \ref{rel2desc} shows the relations among entities and their corresponding synonyms. The relation $\mathrm{identical}$ links a entity and a synonym to claim that the synonym refers to the entity. During training, the relation $\mathrm{identical}$ links $\M$ and a synonym or entity to extract the textual feature. Among other relations, $\mathrm{is\_a}$ is the most common relation, which describes the subsumption relation between a child entity and a parent entity. We transform these relations into phrases to put them in templates used by our Prompt-based model.

\begin{table*}[bp]
\caption{\label{rel2desc}Relations and phrases}
\centering
\begin{tabular}{|l|l|c|c|c|c|c|}

\hline Relation                            & Phrase                            & cl & fbbt & doid & mp & hp \\
\hline identical                           & is identical with                 &  \checkmark  &  \checkmark    &  \checkmark    &  \checkmark  &  \checkmark  \\
\hline is\_a                               & is a kind of                      &  \checkmark  &  \checkmark    &  \checkmark    &  \checkmark  &  \checkmark  \\
\hline capable\_of                         & is capable of                     &  \checkmark  &      &      &    &    \\
\hline negatively\_regulates               & negatively regulates              &  \checkmark  &      &      &    &    \\
\hline positively\_regulates               & positively regulates              &  \checkmark  &      &      &    &    \\
\hline regulates                           & regulates                         &  \checkmark  &      &      &    &    \\
\hline part\_of                            & is part of                        &  \checkmark  &  \checkmark    &      &    &    \\
\hline has\_part                           & has                               &  \checkmark  &      &      &    &    \\
\hline develops\_from                      & develops from                     &  \checkmark  &  \checkmark    &      &    &    \\
\hline has\_sensory\_dendrite\_in          & has sensory dendrite in           &    &  \checkmark    &      &    &    \\
\hline sends\_synaptic\_output\_to         & sends synaptic output to          &    &  \checkmark    &      &    &    \\
\hline synapsed\_to                        & is synapsed to                    &    &  \checkmark    &      &    &    \\
\hline synapsed\_by                        & is synapsed by                    &    &  \checkmark    &      &    &    \\
\hline continuous\_with                    & is continuous with                &    &  \checkmark    &      &    &    \\
\hline synapsed\_via\_type\_Ib\_bouton\_to & is synapsed via type Ib bouton to &    &  \checkmark    &      &    &    \\
\hline receives\_synaptic\_input\_in       & receives synaptic input in        &    &  \checkmark    &      &    &    \\
\hline overlaps                            & overlaps                          &    &  \checkmark    &      &    &   \\
\hline
\end{tabular}
\end{table*}

\subsection{Implementation details}
\label{subsec:appendixA2}
\textbf{Details about prompt-based methods}
For prompt-based methods (Prompt, GraphPrompt (w/o $\Temp^2$), and GraphPrompt), we trained the model with $\mathcal{L}_\mathrm{c}$ for 400 iterations to warm-up entity embedding $\uu_v$. For zero-shot setting, we followed the bi-encoder architecture that uses two encoders for entities and synonyms. Every time we updated the embeddings of entities $\uu_{v'}, v'\in \mathcal{V}$, we had to run the encoder for every entity. For few-shot learning, we found that the entity embedding can be directly trained with an embedding layer. We used the entity side of the bi-encoder to generate entity embedding $\uu_{v'}^0$, and used this embedding to initialize the embedding layer. Then we used embeddings from this trainable embedding layer to replace the $\sg(\uu_{v'})$ and $\sg(\uu_{v})$ term in the loss. 

\textbf{Details about second-order GraphPrompt}
The second-order GraphPrompt (GraphPrompt in \textbf{Table 1}) actually didn't include zeroth-order and first-order templates, since we considered that they are sub-templates of second-order templates. We achieved this by padding a $\MASK$ neighbor. For example, $\Temp^0(\M, v)$ is implemented as $\Temp_\mathrm{identical}^2(\M, v, \M)$, and $\Temp_r^1(\M, v)$ is implemented as $\Temp_r^2(\M, v, \M)$. To get $\uu_v$ and $\vv_v$ from this template, you only need to ignore the output of the second mask.

\textbf{Details about the base model}
The base model is a BioSyn-like model with some important modifications. We trained the model for 30 epochs with initial learning rate 1e-5, and decayed it to about 1e-6 when the model converged. We used sparse features \citep{sung2020biomedical} during candidate generation. During encoding, we didn't add sparse features, since we found that sparse features had no significant impact on the results, and even caused a slight decrease of accuracy. Besides, we found that BioSyn \citep{sung2020biomedical} sometimes failed to retrieve positive candidates due to limited candidate size and the inaccuracy of the model. Therefore, we manually added positive candidate in order to make full use of training data. The procedure for inference is the same as BioSyn \citep{sung2020biomedical}.

\textbf{Details about baselines}
NormCo \citep{wright2019normco} was initially introduced to perform bio-entity linking with inputs being text corpora. Central to its proposed method is the modeling of coherence leveraging concept co-mentions in each text corpus. However, as NormCo is not designed to learn the semantics of concepts, it is not capable of zero-shot learning in our dataset. Therefore, we did not report its results under zero-shot setting. 

In addition, to construct a coherence sequence analogous to the co-appearances of mentions in the original setting, we took the mentions (entities and synonyms) of neighbor concepts of each training concept (excluding validation and test mentions when training), where the mentions are arranged in order based on their distance from the central concept we build this sequence for.

\par
As Sieve-Based \citep{d2015sieve} is a rule-based entity normalization method which does not need the training data, we treated the model as a zero-shot model. Besides, Sieve-Based does not include a scoring mechanism, so we could only report the results of Acc@1. 

\bibliography{aaai23}
\bibliographystyle{aaai23}